\documentclass[10pt,twocolumn,letterpaper]{article}

\usepackage{cvpr}      
\usepackage{multirow}
\usepackage{pifont}
\usepackage{amssymb}
\usepackage{booktabs}
\usepackage{relsize}
\usepackage{amsmath}
\usepackage{graphicx}
\definecolor{cvprblue}{rgb}{0.21,0.49,0.74}
\usepackage[pagebackref,breaklinks,colorlinks,allcolors=cvprblue]{hyperref}

\def\paperID{10293} 
\def\confName{CVPR}
\def\confYear{2026}

\title{TALENT: Target-aware Efficient Tuning for Referring Image Segmentation}

\author{
Shuo Jin\textsuperscript{1,2}
\quad
Siyue Yu\textsuperscript{1}\thanks{Corresponding author: siyue.yu02@xjtlu.edu.cn}
\quad
Bingfeng Zhang\textsuperscript{3}
\quad
Chao Yao\textsuperscript{4}
\quad
Meiqin Liu\textsuperscript{5}
\quad
Jimin Xiao\textsuperscript{1}
\quad
\\
\textsuperscript{1}XJTLU
\quad
\textsuperscript{2}University of Liverpool
\quad
\textsuperscript{3}China University of Petroleum (East China)
\\
\textsuperscript{4}University of Science and Technology Beijing
\quad
\textsuperscript{5}Beijing Jiaotong University
\\
{\tt\small shuo.jin@liverpool.ac.uk},
{\tt\small \{siyue.yu02, jimin.xiao\}@xjtlu.edu.cn},
\\
{\tt\small bingfeng.zhang@upc.edu.cn},
{\tt\small yaochao@ustb.edu.cn},
{\tt\small mqliu@bjtu.edu.cn}
}

\begin{document}
\maketitle
\begin{abstract}
Referring image segmentation aims to segment specific targets based on a natural text expression. Recently, parameter-efficient tuning (PET) has emerged as a promising paradigm. However, existing PET-based methods often suffer from the fact that visual features can't emphasize the text-referred target instance but activate co-category yet unrelated objects. We analyze and quantify this problem, terming it the `non-target activation' (NTA) issue. To address this, we propose a novel framework, TALENT, which utilizes target-aware efficient tuning for PET-based RIS. Specifically, we first propose a Rectified Cost Aggregator (RCA) to efficiently aggregate text-referred features. Then, to calibrate `NTA' into accurate target activation, we adopt a Target-aware Learning Mechanism (TLM), including contextual pairwise consistency learning and target centric contrastive learning. The former uses the sentence-level text feature to achieve a holistic understanding of the referent and constructs a text-referred affinity map to optimize the semantic association of visual features. The latter further enhances target localization to discover the distinct instance while suppressing associations with other unrelated ones. The two objectives work in concert and address `NTA' effectively. Extensive evaluations show that TALENT outperforms existing methods across various metrics (\eg, 2.5\% mIoU gains on G-Ref val set). Our codes will be released at: \url{https://github.com/Kimsure/TALENT}.
\end{abstract}    
\section{Introduction}
\label{sec:intro}

\begin{figure}[!t]
\centering
\includegraphics[width=0.47\textwidth]{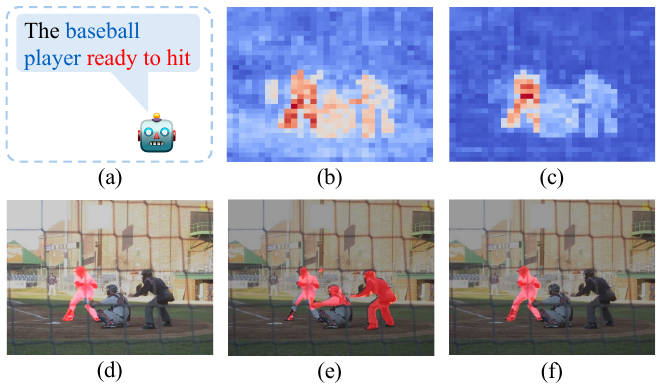}
\caption{Visual feature activation and segmentation maps. (a) Text descriptions. (b) Visual-text fusion in DETRIS~\cite{huang2025densely}, which activates co-category foreground objects. (c) Our TALENT, which emphasizes the text-referred target instance. (d) Segmentation result of the ground truth (GT). (e)-(f) Corresponding segmentation results of DETRIS~\cite{huang2025densely} and our TALENT.}
\label{fig1}
\end{figure}

Referring image segmentation (RIS) aims to segment a specific object within an image guided by a natural language expression, \ie,  RIS requires a precise visual-text alignment to build a `one-to-one' correspondence between the textual expression and the visual regions. This demands fine-grained alignment across diverse object categories, attributes, and spatial relations, making RIS one of the most challenging tasks in vision-language understanding.

Early studies~\cite{lavt,yang2024remamber,cris,hu2023onetoonerethinkingreferringimage,Huang_Fu_Liu_Jiang_Yu_Song_2025,pan2024rethinking} have demonstrated the effectiveness of parameter-full tuning (PFT), which tunes full parameters in powerful models~\cite{swin,resnet}. Yet, this paradigm introduces a significant training computational overhead as the model size scales up~\cite{fang2023eva,he2022masked,oquab2023dinov2,tang2025seeing,tang2025intervening,xue2025mmrc,yang2025streamagent}.
To reduce the training burden, parameter-efficient tuning (PET) methods are proposed, where image-text adapters~\cite{gao2024clipadapter,zhou2022CoOp,wang2025FGPT,wang2025normal} are used to jointly finetune the vision and text features extracted from the frozen pretrained backbones for final text-referred segmentation masks. 

However, current PET-based frameworks tend to focus on salient, semantically equivalent objects, thereby neglecting the specific target instances designated by the text expressions. As illustrated in Fig.~\ref{fig1}, given the textual query “the baseball player ready to hit”, the existing method fails to accurately identify the intended target instance (“ready to hit”). Instead, it is misled by the visual prominence of homogeneous objects of the same category (“baseball player”), as shown in Fig.~\ref{fig1}~(b). 
We term this issue `non-target activation (NTA)', which ultimately leads to an incorrect segmentation result, as depicted in Fig.~\ref{fig1}(e).

\begin{figure}[!t]
\centering
\includegraphics[width=0.43\textwidth]{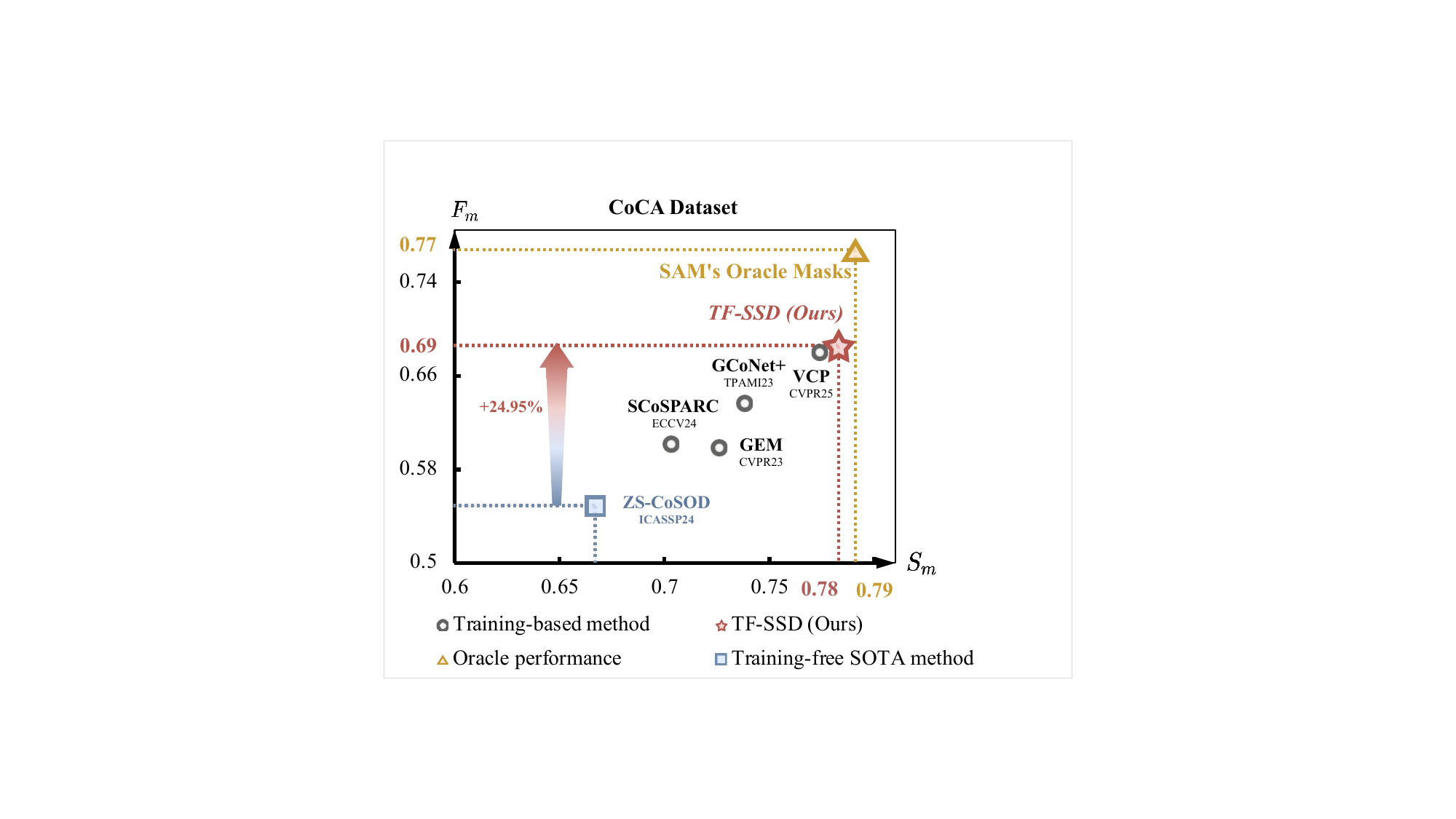}
\caption{Comparison of `NTA' influence quantization among various methods on the selected subset of different benchmarks.}
\label{fig-amiou}
\end{figure}

Further, to quantify the impact of `NTA', we calculate the proportion of incorrectly predicted regions that fall into other actual non-target instances within the same category. This proportion is named as NTA-IoU to evaluate how strongly the `NTA' issue affects PET-based methods~\cite{wang2023barleria,yu2024etog,huang2025densely,peng2025parameter,yang2025ffr}. Note that a higher NTA-IoU score indicates a stronger influence of `NTA', whereas a lower score suggests a weaker `NTA' effect and, therefore, a better segmentation outcome. Fig.~\ref {fig-amiou} reports the NTA-IoU of different methods. Obviously, existing PET-based approaches exhibit higher NTA-IoU scores, where the model tends to segment salient co-category objects rather than the specific text-described target instance.

As described above, our goal is to mitigate the `NTA' issue to achieve precise target activation and yield more accurate target segmentation. Accordingly, we propose a \textbf{T}arget-\textbf{A}ware \textbf{L}earning framework to suppress `NTA' for \textbf{E}fficie\textbf{N}t \textbf{T}uning RIS, termed \textbf{TALENT}. First, we design a Rectified Cost Aggregator (RCA), which constructs a cost volume to represent the visual-text relationship for text-referred visual feature aggregation. Then, we introduce the Target-aware Learning Mechanism (TLM) to optimize the aggregated features from RCA and thus calibrate the `NTA' issue, which includes Contextual Pairwise Consistency Learning (CPCL) and Target Centric Contrastive Learning (TCCL).

In detail, CPCL provides a holistic understanding of the referring text by leveraging global text features, along with aggregated RCA features, to construct a text-augmented pairwise feature affinity map. This map guides the model in refining the original RCA feature relationships and learning context-aware semantic associations. However, although the global text feature provides a holistic understanding of one object, it exhibits coarse granularity, causing CPCL to overlook certain fine-grained semantic details. To further enhance the discrimination of the distinct target instance, we introduce TCCL, which strengthens the alignment between visual features and the corresponding target text expression while simultaneously reducing their association with other non-target textual descriptions. 
These two learning objectives work in concert to enable our target-aware learning, allowing our model to effectively improve feature discriminability and thereby alleviate the `NTA' issue. As depicted in Fig.~\ref{fig1}~(c), TALENT helps the fused feature emphasize the text-referred object and segments the precise instance shown in Fig.~\ref{fig1}~(f), which closely aligns with the GT segmentation result in Fig.~\ref{fig1}~(d). In addition, TALENT achieves up to triple gains of the NTA-IoU score compared to other PET-based methods, as shown in Fig.~\ref{fig-amiou}.

Comprehensive experiments demonstrate that our TALENT outperforms existing approaches, including both PFT-based and PET-based methods, across various benchmarks. In summary, our main contributions are as follows:

\begin{itemize}
\item We identify and quantify the `NTA' issue in PET-based RIS, where visual features attend to salient objects instead of the specific target instance. To address this, we propose TALENT for PET-based RIS through target-aware learning mechanism, where a Rectified Cost Aggregator (RCA) is first introduced for visual-text aggregation. 

\item We design Target-aware Learning Mechanism (TLM), including Contextual Pairwise Consistency Learning (CPCL) and Target Centric Contrastive Learning (TCCL), to suppress `NTA': CPCL constrains the RCA feature to learn context-aware semantic relations and TCCL forces it to identify the distinct target instance.

\item We conduct comprehensive experiments to demonstrate that TALENT achieves state-of-the-art (SOTA) performance across various benchmarks. (\eg, TALENT achieves a gain of 1.9\% mIoU and 3.3\% oIoU on the RefCOCO+ testB set over PET-based methods).
\end{itemize}

\section{Related Work}
\label{sec:formatting}
Our work introduces a parameter-efficient tuning approach for RIS. In this section, we summarize two mainstream paradigms and discuss their differences. 
\begin{figure*}[!t]
\centering
\includegraphics[width=0.95\textwidth]{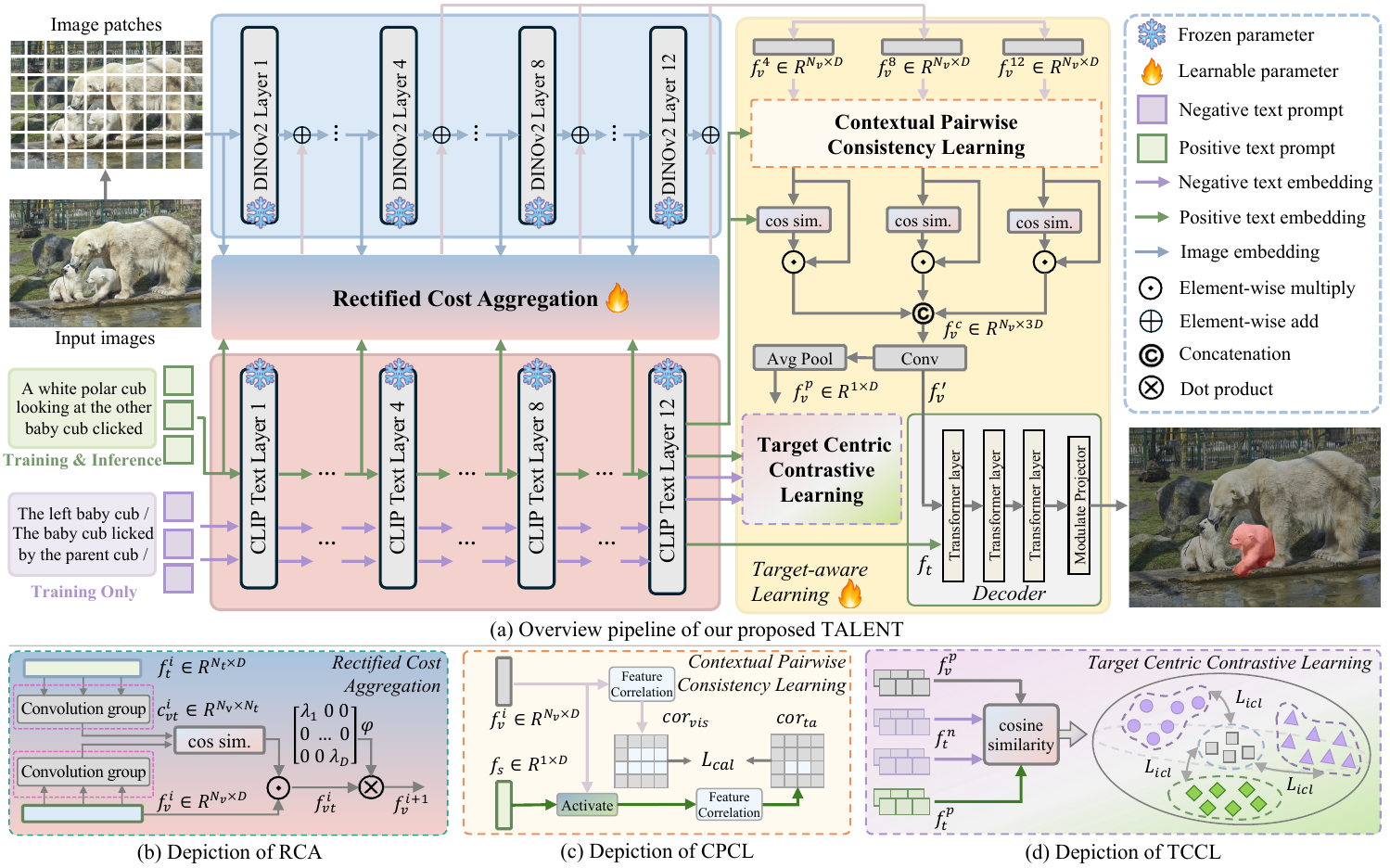}
\caption{Framework pipeline of our TALENT. It contains four main modules: a frozen backbone building upon DINOv2-Reg and CLIP to encode the image and text, a rectified cost aggregator for vision-language interaction, a target-aware learning mechanism to strengthen feature representation, and a transformer decoder for final segmentation.}
\label{fig2}
\end{figure*}

\noindent
\textbf{Parameter-full Tuning for RIS.} RIS is originally introduced by~\cite{RIS1}, which aims to predict a pixel-level object mask described by a text expression. Early methods~\cite{mcn,cmpc+,brinet,li2023fully,sun2021discriminative,sun2021iterative} typically encode the referring expression into a fixed-length feature vector that is then concatenated with the visual feature. It enabled multimodal interaction through convolutional frameworks like FCN~\cite{fcn}. Subsequent methods~\cite{vlt,lavt,liu2023caris} have advanced this paradigm by incorporating shallow visual and text features to perform cross-modal fusion at multiple feature levels, which can preserve coarse-to-fine spatial details. Some methods~\cite{cheng2025weakmcn,dai2025multi,liu2022temporal,jin2023kernel} also explore the potential of multi-task learning to obtain fine-grained results, where the bounding box of the referring expression can be treated as pseudo labels to enhance the referring segmentation. In addition, multimodal pretraining introduces the advantage of large-scale data. CRIS~\cite{cris} emphasizes sentence-pixel alignment to leverage multimodal information, while PCAN~\cite{chen2022position} focuses on position-aware contrastive alignment. However, these methods require PFT, incuring extra computation and limited scalability due to the training burden.

\noindent
\textbf{Parameter-efficient Tuning for RIS.} PET adjusts only a fraction of the parameters and alleviates the computational burden compared to PFT-based methods. Prominent approaches~\cite{gao2024clipadapter,jie2022convolutional} are designed to integrate lightweight modules into the frozen backbone~\cite{clip,jin2025feature,qiu2025bias}, where only the extra components are updated during tuning. It enables a residual tuning pipeline. Another paradigm for PET is to update only a subset of the original model parameters. Some researchers have explored matrix decomposition methods~\cite{hu2021lora} to reduce learnable parameters by factorizing the weights of pretrained models. Recently, PET-based approaches for RIS have been promisingly improved. ETRIS~\cite{xu2023bridging} introduces PET to referring image segmentation by leveraging the bridge module for visual-text fusion. BarLeRIa~\cite{wang2023barleria} builds an intertwined adapter along with a normalizing flow mechanism with extremely few trainable parameters. DETRIS~\cite{huang2025densely} employs multiscale convolutional adapters on both vision and text backbones to achieve a significant performance gain. Nevertheless, these PET-based methods don't thoroughly examine whether visual features can effectively respond to text-referred visual regions, \eg, whether the `NTA' issue can be solved. In contrast, we conduct an in-depth analysis of the `NTA' issue and propose an effective PET-based solution for RIS. 

\section{method}
\subsection{Problem Quantization}

As analyzed in Sec.~\ref{sec:intro}, the `NTA' issue arises from the fact that existing approaches tend to emphasize salient objects within the input image, instead of the text-described specific target instance.

To quantify this issue, we introduce the NTA-IoU metric to evaluate whether the model can accurately distinguish the correct instance from salient objects of the co-category. Specifically, we select a subset of widely used benchmarks: RefCOCO, RefCOCO+, and G-Ref, which satisfy the following criteria: each image contains multiple text expressions referring to different co-category instances. We define the union of all these GT masks $\{G_1, \dots, G_n\}$ as the co-category GT $G$. For a given segmentation map $I_1$ corresponding to the text-referred object, we exclude the overlap with its matched GT mask $G_1$, and then compute the IoU between the remaining predicted region and the co-category GT mask. Our NTA-IoU metric can be formulated as:
\begin{equation}
    \text{NTA-IoU}=\frac{(I_1-I_1 \cap G_1) \cap (G-G_1)}{(I_1-I_1 \cap G_1) \cup (G-G_1)}.
\end{equation}
Intuitively, NTA-IoU measures the proportion of the incorrectly segmented regions that fall into non-target instances within the same category. A higher NTA-IoU indicates that the model struggles to distinguish between the intended target instance and its non-target objects of the same category, and vice versa.

\subsection{Overview}
Fig.~\ref{fig2}~(a) depicts the overview of our framework, including several modules: a frozen backbone building on DINOv2-Reg~\cite{oquab2023dinov2} and CLIP~\cite{radford2021learning}, a Rectified Cost Aggregator (RCA), a Target-aware Learning Mechanism (TLM), including Contextual Pairwise Consistency Learning (CPCL) and Target Centric Contrastive Learning (TCCL), and a transformer decoder. The complete training process is:

\begin{enumerate}
\item First, the image is input to the frozen DINOv2-Reg encoder for visual features. Both positive and negative texts are input to the frozen CLIP text encoder for text features. Note that the negative texts are only for TCCL.
\item Instead of directly propagating visual features through DINOv2-Reg layers, the intermediate feature is fused with the corresponding positive text feature through RCA for efficient tuning and then input to the next layer.
\item Then, the visual features from RCA across different layers participate in our CPCL, where the positive sentence-level text feature $f_s$ is utilized to construct a text-augmented pairwise feature affinity map to strengthen context-aware semantic association of visual features.
\item After that, $f_s$ guides each visual feature via a cosine similarity mask to strengthen the text-referred features. Then, the strengthened multi-layer features are aggregated into a single global feature for our TCCL. It further enhances feature discrimination to identify the target instance by leveraging both positive and negative texts.
\item Finally, a transformer decoder generates the final segmentation map. During inference, only the given positive text prompt is used with the image for prediction. 
\end{enumerate}

\subsection{Rectified Cost Aggregator} 
As mentioned above, existing PET-based methods can't emphasize the precise target activation. Enlightened by CAT-Seg~\cite{cho2024cat},
we propose a Rectified Cost Aggregator (RCA) for visual-text interaction. By constructing vectorized cost-volume, we rectify the model’s focus on text-referred regions by suppressing adverse cross-modal interactions.

First, or a tokenized positive text expression $T \in \mathbb{R}^{L}$ and an input image $I \in \mathbb{R}^{H \times W \times C}$, the frozen CLIP text encoder and DINOv2-Reg encoder are deployed to extract corresponding vectorized features with a fixed token length~\cite{radford2021learning}. Then, our RCA projects them into the same channel dimension. Suppose the projected text feature $f_t^{i} \in \mathbb{R}^{N_t \times D}$ and the visual feature $f_v^{i} \in \mathbb{R}^{N_v \times D}$ are the output features of the $i$-th layer, where $N_t$ and $N_v$ denote the token length of the text and visual features, respectively. Then, RCA uses parallel convolutions to project visual and text features $f_v^{i}$ and $f_t^{i}$ into the $\hat{f}_v^i$ and $\hat{f}_t^i$.

Next, cost volume is utilized to localize referring regions through matching text and visual features~\cite{cho2024cat}, which is shown in Fig.~\ref{fig2}~(b). To suppress irrelevant visual-text interactions in the vectorized cost space, the ReLU activation function is applied to filter out negative matching responses and rectify the positive alignment.
It is formulated as,
\begin{equation}
c_{vt}^i = \text{ReLU}(\frac{\hat{f}_v^i \cdot \hat{f}_t^i}{\|\hat{f}_v^i\| \cdot \|\hat{f}_t^i\|}),
\end{equation}
where $c_{vt}^i \in \mathbb{R}^{N_v \times N_t}$ denotes the matching cost volume at the $i$-th layer. It can be treated as a text-referred semantic mask to guide the visual feature as follows,
\begin{equation}
f_{vt}^{i} =P_{ex}^i(c_{vt}^i)\odot f_{v}^i,
\end{equation}
where $P_{ex}(\cdot)^i$ denotes a linear layer to align the dimension of $c_{vt}^i$  and $f_{v}^i$ , $\odot$ denotes the element-wise multiplication.

To minimize disruption to the feature extraction capability of the frozen vision backbone, we further apply a learnable diagonal matrix~\cite{touvron2021going} to rescale the fused feature as a residual, which is then injected into the next layer:
\begin{equation}
f_{v}^{i+1} = \nu_{i+1}(f_{v}^{i}) + f_{vt}^{i}*\varphi,
\label{eq:4}
\end{equation}
where $\nu_i(\cdot)$ denotes the $i$-th layer of the vision backbone, $\varphi=\text{diag}(\lambda_1, . . . , \lambda_D)$ denotes the rescaling diagonal matrix. In this way, the visual feature is guided by the text expression and propagated during the image encoder to generate the text-referred visual feature.

\subsection{Target-aware Learning Mechanism}
Although our RCA generates text-referred visual features, the `NTA' issue still persists. It often arises with semantic ambiguity, where the model tends to activate salient co-category objects rather than the distinct text-described target instance. Thus, we propose calibrating `NTA' through a Target-aware Learning Mechanism (TLM), with two learning objectives: 1) strengthen the context-aware semantic relations of visual features through Contextual Pairwise Consistency Learning (CPCL); 2) facilitate feature discrimination to identify the target instance through Target Centric Contrastive Learning (TCCL).

\noindent
\textbf{Contextual Pairwise Consistency Learning.}
In CPCL, we aim at constructing the text-augmented semantic relationships to refine multi-layer visual features $f_v^i$ from RCA, where $i\in[4,8,12]$. 
As shown in Fig.~\ref{fig2}~(c). CPCL applies the holistic, sentence-level text feature $f_s \in \mathbb{R}^{1 \times D}$, which is defined as the [EOS] token of the CLIP text encoder~\cite{radford2021learning}, to activate the visual feature and obtain the text-augmented feature $f_{ta}^i \in \mathbb{R}^{N_v \times 1}$, formulated as,
\begin{equation}
\begin{aligned}
    f_{ta}^i = f_v^i \cdot f_s^\top,
\end{aligned}
\end{equation}
where $\top$ denotes the transpose operation. Thereby, $f_{ta}^i$ contains a global coherence between the sentence-level text token and each visual token.

Then, to make visual features learn context-aware semantic relations that are associated with the textual description, we derive the visual and text-augmented feature correlation maps $cor_{vis}^i$ and $cor_{ta}^i$ as follows,
\begin{equation}
\begin{aligned}
cor_{vis}^i &= \mathrm{Softmax} \left( \frac{f_v^i \cdot (f_v^i)^\top}{\|f_v^i\|^2} \right), \\
cor_{ta}^i &= \mathrm{Softmax} \left( \frac{f_{ta}^i \cdot (f_{ta}^i)^\top}{\|f_{ta}^i\|^2} \right).
\end{aligned}
\end{equation}
Here, $cor_{vis}^i$ and $cor_{ta}^i$ represent their pair-wise feature relationship of the $i$-th layer. Since $f_{ta}^i$ is strengthened by text feature, $cor_{ta}^i$ has a strong text-augmented feature relationship. To address `NTA', we aim to align $cor_{vis}^i$ with $cor_{ta}^i$, which helps $f_v^i$ accurately activate the plausible visual regions. Hence, we minimize the cosine distance to optimize the feature correlation $cor_{vis}^i$, where the optimization area is constrained by $cor_{ta}^i$. In this way, the visual feature enhances the accurate target activation in text-referred regions. This consistency loss $\mathcal{L}_{cpcl}$ is formulated as:
\begin{equation}
    \mathcal{L}_{cpcl}=\sum_{i\in[4,8,12]}\| (\boldsymbol{J} - cor_{vis}^i) \odot cor_{ta}^i\|^2_F,
\label{eq7}
\end{equation}
where $\boldsymbol{J}$ denotes the all-ones matrix, and $\| \cdot \|_{F}^2$ denotes the Frobenius norm. Note that $cor_{ta}^i$ is detached as a pseudo label. CPCL optimizes visual features $f_v^i$ to exhibit context-aware feature relations, and enhances the visual-text semantic coherence for target region perception.

Next, $f_v^i$ are multiplied with $f_s$ via cosine similarity maps, which are used to reweight $f_v^i$ for interaction. Finally, they are concatenated as $f_v^c \in \mathbb{R}^{N_v \times 3D}$ for our TCCL.

\noindent
\textbf{Target Centric Contrastive Learning.}
CPCL constrains the visual feature to highlight the correct context region by enhancing semantic associations. However, the enhanced feature tends to exhibit coarse representation due to its optimization with holistic semantic understanding, where the model primarily attends to high-level semantics rather than instance-specific details. 
To identify the specific instance for fine-grained target perception, 
we propose TCCL to enhance the feature discrimination by drawing the positive text features closer to the visual representation while repelling negative ones belonging to irrelevant instances. Fig.~\ref{fig2}~(d) depicts the detailed processing of our TCCL.

Firstly, to construct instance-level positive and negative pairs, we use the given text that describes the target as the positive sample and collect those referring to another object in the same image as negative samples, as shown on the left of Fig.~\ref{fig2}~(a). Then, these text prompts are input to the frozen CLIP text encoder for positive and negative text features, $f_s^p$ and $f_s^n$, respectively. 

Synchronously, we obtain a global prototype $f_v^p \in \mathbb{R}^{1 \times D}$ from the fused multi-layer feature $f_v^c$ to align with the sentence-level text features, formulated as,
\begin{equation}
    f_v^p = \text{Avg}(\text{conv}(f_v^c)),
\end{equation}
where $\text{Avg}(\cdot)$ denotes the average pooling operation and $\text{conv}(\cdot)$ denotes a convolution layer to reduce channels. Then, text features are fused with $f_v^p$ via the cosine similarity for contrastive learning. Our $\mathcal{L}_{tccl}$ aims to optimize distances among various features, which is formulated as,
\begin{equation}
\scalebox{1.05}{$
\mathcal{L}_{tccl} = -\log \frac{\exp(\text{sim}(f_v^p, f_s^p))}{\exp(\text{sim}(f_v^p, f_s^p)) + \sum\limits_{k=1}^{K} \exp(\text{sim}(f_v^p, f_s^{n_k}))},
$}
\end{equation}
where $\text{exp}(\text{sim}(\cdot, \cdot))$ denotes exponential of cosine similarity, and $K$ denotes the number of negative samples. In this way, $\mathcal{L}_{tccl}$ enhances the linkage between visual feature and its textual description for accurate target perception.

Overall, $\mathcal{L}_{cpcl}$ and $\mathcal{L}_{tccl}$ collaboratively support our target-aware learning. $\mathcal{L}_{cpcl}$ refines the pairwise semantic association of visual features, offering a plausible yet coarse target region, while $\mathcal{L}_{tccl}$ further enhances the feature discrimination for fine-grained target instance perception.
This joint optimization of TLM enables our model to focus on the text-referred target and mitigate the `NTA' issue. 
\begin{table*}[t]
\centering
\small
\setlength\tabcolsep{8.0pt}
\caption{Quantitative comparison of our method against other PFT-based and PET-based methods on different benchmarks, evaluated using the oIoU and mIoU metrics. $^\dagger$ denotes the re-implemented results, while others are directly sourced from the reported results. u: The UMD partition. g: The Google partition. The best results are marked with \textbf{bold}.}
\renewcommand{\arraystretch}{0.9}
\begin{tabular}{l|c|c|ccccccccc}
\toprule[0.3mm]
\multirow{2}{*}{Methods}   & \multirow{2}{*}{PET}  & \multirow{2}{*}{Metirc}      & \multicolumn{3}{c|}{RefCOCO}       & \multicolumn{3}{c|}{RefCOCO+}                              & \multicolumn{3}{c}{G-Ref}     \\ \cline{4-12}
          &     &        & \multicolumn{1}{c|}{val} & \multicolumn{1}{c|}{testA} & \multicolumn{1}{c|}{testB} & \multicolumn{1}{c|}{val} & \multicolumn{1}{c|}{testA} & \multicolumn{1}{c|}{testB} & \multicolumn{1}{c|}{val(u)} & \multicolumn{1}{c|}{test(u)} & \multicolumn{1}{c}{val(g)}                       \\ \midrule
LAVT$^\dagger$~\cite{lavt}  &  \ding{56}   & \multirow{8}{*}{oIoU}                   & 72.7                     & 75.8                       & 68.8                       & 62.1                     & 68.4                       & 55.1                       & 61.2                        & 62.1                         & 60.5                                            \\
DMMI~\cite{hu2023onetoonerethinkingreferringimage}       &  \ding{56}      &                    & 74.1                     & 77.1                       & 70.2                       & 64.0                     & 69.7                       & 57.0                       & 63.5                        & 64.2                         & 61.2                                           \\
ReLA~\cite{gres}       &  \ding{56}      &                    & 73.8                     & 76.5                       & 70.2                       & 66.0                     & 71.0                       & 57.7                       & 65.0                        & 66.0                         & 62.7                                            \\
CG-former~\cite{tang2023contrastive}        &  \ding{56}   &                 & 74.8                     & 77.3                       & 70.6                       & 64.5                     & 71.0                       & 57.1                       & 64.7                        & 65.1                         & 62.5                                          \\
LISA-Vicuna-13B~\cite{lai2024lisa}       &  \ding{56}   &                   & 71.7                     & 74.7                       & 68.1                       & 59.4                     & 64.2                       & 52.9                       & 65.2                        & 66.1                         & -                                           \\
MagNet~\cite{chng2023mask}       &   \ding{56}  &                    & 75.2                     & 78.2                       & 71.1                       & 66.2                     & 71.3                       & 58.1                       & 65.4                        & 66.2                         & 63.1                                          \\
LQMFormer~\cite{shah2024lqmformer}      &  \ding{56}    &                 & 74.2                     & 76.8                       & 71.0                       & 65.9                     & 71.8                       & 57.6                      & 64.7                        & 66.0                         & 63.0                                            \\
ReMamber~\cite{yang2024remamber}       &  \ding{56}   &                  & 74.5                     & 76.7                       & 70.9                       & 65.0                     & 70.8                       & 57.5                       & 63.9                        & 64.0                         & -                                           \\ 
CoHD~\cite{luo2024cohd} &   \ding{56}     &              & 75.1                    & 78.3                       & 71.0                       & 66.8                     & 71.6                       & 58.4                       & 65.8                        & 66.7                         &  -                                  \\
RISCLIP~\cite{kim2024extendingclipsimagetextalignment}       &  \ding{52}     &              & 73.6                  & 76.5                       & 69.8                       & 65.5                     & 70.6                       & 55.5                       & 64.1                        & 65.1                         & -                          \\
ETOG~\cite{yu2024etog}      &   \ding{52}     &              & 71.4                    & 76.1                       & 66.7                       & 62.3                     & 68.5                       & 51.9                       & 61.1                        & 62.8                         &  -                                  \\
TALENT (Ours)          &  \ding{52}        &           & \textbf{75.9}                     & \textbf{78.3}                       & \textbf{72.8}                       &  \textbf{66.9}                        &  \textbf{72.3}                          &  \textbf{58.8}                          &  \textbf{65.9}                          &   \textbf{66.8}                          &  \textbf{64.7}                           \\
\midrule
ETRIS~\cite{xu2023bridging}       & \ding{52}    & \multirow{5}{*}{mIoU}                 & 70.5                     & 73.5                       & 66.6                       & 60.1                     & 66.9                       & 50.2                       & 59.8                        & 59.9                         & 57.9                                    \\
BarLeRIa~\cite{wang2023barleria}        &     \ding{52}      &           & 72.4                     & 75.9                       & 68.3                       & 65.0                     & 70.8                       & 56.9                       & 63.4                        & 63.8                         & 61.6                                        \\
RISCLIP~\cite{kim2024extendingclipsimagetextalignment}       &  \ding{52}     &              & 75.7                   & 78.0                       & 72.5                       & 69.2                     & 73.5                       & 60.7                       & 67.6                        & 68.0                         & -                          \\
ETOG~\cite{yu2024etog}      &   \ding{52}     &              & 73.4                    & 76.9                       & 69.3                       & 66.0                     & 71.5                       & 56.9                       & 63.8                        & 64.6                         &  -                                  \\
DETRIS~\cite{huang2025densely}       &  \ding{52}  &                   & 76.0                     & 78.2                      & 73.5                       & 68.9                     & 74.0                       & 61.5                       & 67.9                        & 68.1                         & 65.9                                  \\
TALENT (Ours)       &   \ding{52}         &           & \textbf{77.8}                     & \textbf{79.4}                       & \textbf{74.8}                       &  \textbf{70.1}                        &  \textbf{74.9}                          &  \textbf{63.4}                          &  \textbf{69.7}                          &   \textbf{69.1}                           &  \textbf{68.4}                                   \\   
\bottomrule[0.3mm]
\end{tabular}
\label{tab1}
\end{table*}

\subsection{Optimization Objective} 
Following previous works~\cite{cris,xu2023bridging,huang2025densely}, we utilize a transformer decoder to convert the derived visual feature from our TLM into the segmentation feature $f_{seg}$. Then, a text-to-pixel discriminative loss~\cite{cris} is deployed to encourage the initial alignment of textual embeddings with the corresponding visual pixels, defined as:
\begin{equation}
\scalebox{0.9}{$
\begin{aligned}
\mathcal{L}_{dis}^j \left( f_{seg}^j, f_t \right) &=
\begin{cases}
    -\log\left( \sigma\left( f_{seg}^j \cdot f_t \right) \right), & j \in \mathcal{P} \\
    -\log\left( 1 - \sigma\left( f_{seg}^j \cdot f_t \right) \right), & j \in \mathcal{N}
\end{cases} \\[1ex]
\mathcal{L}_{dis} \left( f_{seg}, f_t \right) &=
\frac{1}{|\mathcal{P} \cup \mathcal{N}|}
\sum_{j \in \mathcal{P} \cup \mathcal{N}} 
\mathcal{L}_{dis}^j \left( f_{seg}^j, f_t \right),
\end{aligned}
$}
\end{equation}
where $\mathcal{P}$ and $\mathcal{N}$ denote the class of `1' and `0' in the ground truth, $\sigma$ is the sigmoid function. 

To train the overall network, the final objective loss is 
\begin{equation}
    \mathcal{L}_{total} = \mathcal{L}_{dis} + \lambda_{cpcl} \mathcal{L}_{cpcl} + \lambda_{tccl} \mathcal{L}_{tccl},
\end{equation}
where $\lambda_{cpcl}$ and $\lambda_{tccl}$ are the scaling coefficients. Our TLM further mitigates the `NTA' issue.

\section{experiment}
\label{sec:rationale}

\subsection{Datasets and Metrics}
\textbf{Datasets and metrics}. We conduct comprehensive experiments on three benchmarks: RefCOCO~\cite{refcoco}, RefCOCO+~\cite{refcoco}, and G-Ref~\cite{grefcoco}, all collected from MS-COCO~\cite{lin2014microsoft}. To unify the evaluation metrics for a fair comparison, we use overall Intersection-over-Union (oIoU), mean Intersection-over-Union (mIoU), and Precision@X to evaluate segmentation results respectively. The Precision@X metric measures the percentage of test images with an IoU score higher than the threshold X $\in$ $\{$0.5, 0.7, 0.9$\}$.

\subsection{Implementation Details}
We build our model using DINOv2-Reg (ViT-B/14) as the vision encoder and CLIP (ViT-B/16) as the text encoder. Our RCA is applied at encoder layers [1,3,5,7,9,11], where the learnable parameter in the rescaling matrix is initialized to 0.2. The coefficients $\lambda_{cpcl}$ and $\lambda_{tccl}$ are both set as 0.1.
We train our network for 50 epochs using the Adam optimizer with a learning rate of 1e-4, where the input image is resized to 448 $\times$ 448. A learning rate decay is employed at the 35$th$ epoch with a decay factor of 0.1. Our model is trained on two RTX4090 GPUs with a batch size of 32.
\begin{table}[!htbp]
\centering
\small
\setlength\tabcolsep{7.5pt}
\caption{Quantitative comparison using the Precision@X metric on the val-test set of RefCOCO.}
\renewcommand{\arraystretch}{0.85}
\begin{tabular}{lccc}
\toprule[0.3mm]
Methods    & Pr@0.5 & Pr@0.7 & Pr@0.9 \\
\midrule
ETRIS~\cite{xu2023bridging}      &  83.4      &  72.7      &   17.4      \\
ETOG~\cite{yu2024etog}       &  83.2      &   73.0     &   26.2      \\
CG-former~\cite{tang2023contrastive}       &  87.2      &   78.7     &   38.8      \\
Prompt-RIS~\cite{shang2024prompt}     &  85.6      &  76.9      &   26.2       \\
DETRIS~\cite{huang2025densely}     &   87.9     &    80.2    &    27.5      \\
TALENT (Ours)     &    \textbf{88.6}    &   \textbf{82.3}    &   \textbf{40.1}     \\
\bottomrule[0.3mm]
\end{tabular}
\label{tab2}
\end{table}

\begin{figure*}[t]
\centering
\includegraphics[width=0.97\textwidth]{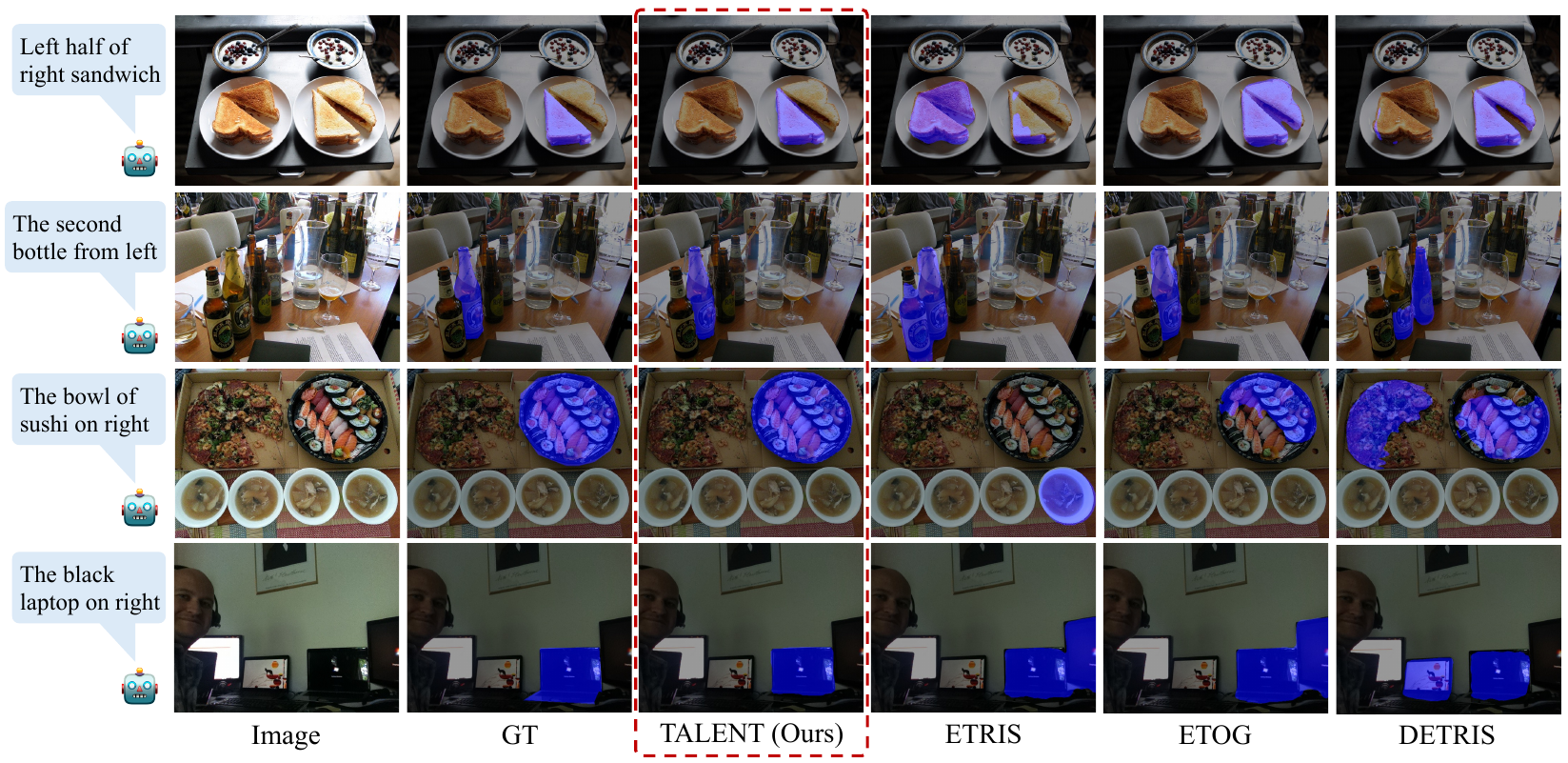}
\caption{Qualitative comparison. We compare our TALENT with ETRIS~\cite{xu2023bridging}, ETOG~\cite{yu2024etog}, and DETRIS~\cite{huang2025densely}. It's observed that TALENT can accurately localize the target and generate more precise segmentation results.}
\label{fig3}
\end{figure*}

\subsection{Quantitative Results}
Our method is evaluated against existing SOTA methods. As shown in Tab.~\ref{tab1}, our method achieves significant gains and sets a new SOTA performance. For example, TALENT outperforms the PET-based method DETRIS~\cite{huang2025densely} by 1.8\% mIoU on the RefCOCO val-test set and also surpasses the PFT-based methods ReMamber~\cite{yang2024remamber} and CoHD~\cite{luo2024cohd}, up to 1.9\% and 1.8\% oIoU on the RefCOCO testB set, with fewer tunable parameters. TALENT even beats LISA-Vicuna-13B~\cite{lai2024lisa}, which requires a large language model, with 3.6\% and 4.7\% oIoU gains on RefCOCO testA and testB sets. These results highlight that our method boosts the segmentation performance of the PET-based paradigm.

Additionally, we evaluate our method against several SOTA methods using the Precision@X metric. Tab.~\ref{tab2} shows that TALENT still outperforms existing methods on the image percentage with different IoU thresholds. Specifically, TALENT outperforms the PET-based method DETRIS~\cite{huang2025densely} 12.6\% and the PFT-based method CG-former~\cite{tang2023contrastive} 1.3\% with the IoU threshold $\text{X}=0.9$. These results demonstrate that our method can precisely segment the accurate targets. It's noted that we report the result of Prompt-RIS~\cite{shang2024prompt} without SAM~\cite{sam} post-processing for a fair comparison.

\subsection{Qualitative Results}
\begin{figure}[!htbp]
\centering
\includegraphics[width=0.48\textwidth]{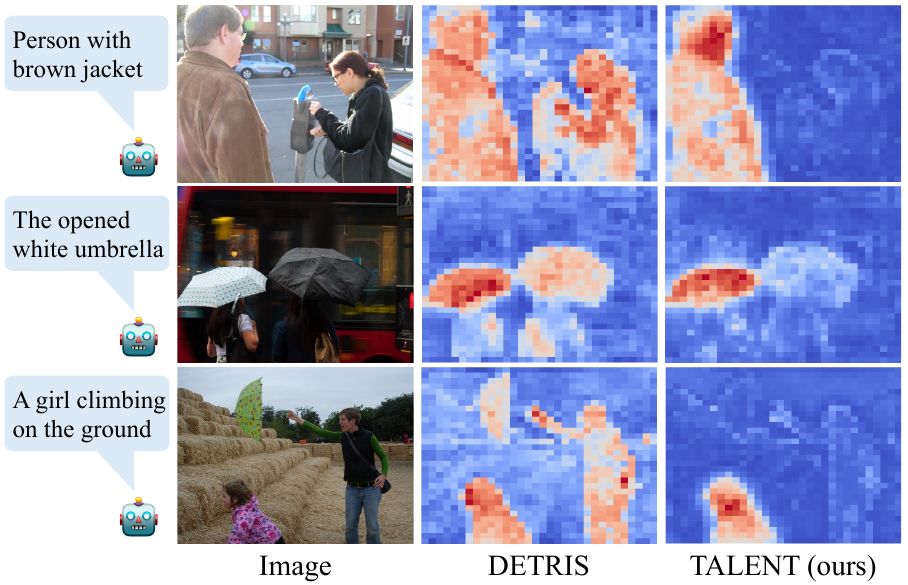}
\caption{Visualization of feature activation maps. We compare TALENT with the existing SOTA method DETRIS~\cite{huang2025densely}.}
\label{fig4}
\end{figure}

In Fig.~\ref{fig3}, we compare segmentation results among different PET-based methods~\cite{xu2023bridging,yu2024etog,huang2025densely}. It's observed that previous methods struggle to localize the text-referred target and often segment other salient and visually similar objects, like `bowl of sushi' in the third row and `black laptop' in the last row. These results indicate that the issue of `NTA' is not well addressed. In contrast, TALENT can precisely localize the text-referred target and predict more accurate segmentation maps, which demonstrates that TALENT effectively mitigates the `NTA' issue. We also visualize the feature activation in Fig.~\ref{fig4}. It's shown that TALENT activates the specific target associated with the text expression, while the previous method often activates salient yet unrelated objects. More feature activation and segmentation results are provided in the appendix.

\subsection{Ablation Study}
\begin{table}[!t]
\centering
\small
\renewcommand{\arraystretch}{0.9}
\setlength\tabcolsep{1.5pt}
\caption{Comparison of parameters and average segment performance on three benchmarks. $^\dagger$ indicates the results are calculated from the official implementation of DETRIS~\cite{huang2020referring} and ETRIS~\cite{xu2023bridging}.}
\begin{tabular}{l|c|c|c|c}
\toprule[0.3mm]
\multirow{2}{*}{Methods} & \multirow{2}{*}{ \begin{tabular}[c]{@{}c@{}}Params.\\ (backbone)\end{tabular}} & \multirow{2}{*}{ \begin{tabular}[c]{@{}c@{}}Params.\\ (other modules)\end{tabular}} & \multirow{2}{*}{ \begin{tabular}[c]{@{}c@{}}Params.\\ (overall)\end{tabular}} & \multirow{2}{*}{ \begin{tabular}[c]{@{}c@{}}mIoU\\ (Avg.)\end{tabular}} \\
    &                      &                          &                            &                   \\ \midrule
CRIS~\cite{cris}     & 57.31M           & 103.94M              & 161.25M            & 63.8           \\
LoRA$^\dagger$~\cite{hu2021lora}           &   \textbf{0.03M}               &  23.98M           &  24.01M         &   60.4     \\
Compacter$^\dagger$~\cite{karimi2021compacter}           &   0.19M               &   23.98M          &   24.17M        &   61.3     \\
ETRIS$^\dagger$~\cite{xu2023bridging}     & 1.94M  & 23.98M              & 25.92M             & 62.8       \\
BarLeRIa~\cite{wang2023barleria}     & 2.21M  & 23.98M              & 26.19M             & 66.5       \\
DETRIS$^\dagger$~\cite{huang2025densely}           & 2.71M                 & 22.84M            & 26.69M          & 70.4       \\
TALENT (ours)       & 2.33M      & \textbf{20.44M}         & \textbf{22.77M}      & \textbf{72.0}                 \\ 
\bottomrule[0.3mm]
\end{tabular}
\label{tab3}
\end{table}

\noindent
\textbf{Evaluation of tunable parameter efficiency.}
We compare our TALENT with adapter-based and decomposition-based PET methods to assess the efficiency of model's tunable parameters. Tab.~\ref{tab3} reports the number of tunable parameters for the backbone and the remaining modules. It's observed that our TALENT achieves the best performance with the fewest extra parameters compared with the adapter-based methods. Furthermore, although several decomposition-based methods, including LoRA~\cite{hu2021lora} and Compacter~\cite{karimi2021compacter}, require fewer parameters to finetune the backbone model, the multimodal interaction between visual and textual features still remains insufficient. In contrast, our method achieves a significant improvement of 10\% mIoU over these decomposition-based methods with the fewest overall tunable parameters. These results demonstrate both the effectiveness and efficiency of our TALENT.

\noindent
\textbf{Effect on various proposed modules.}
Tab.~\ref{tab4} illustrates comprehensive experiments on each proposed module of our TALENT: RCA, CPCL, and TCCL by adding each one incrementally. The baseline model is constructed by simply combining the vision and text backbones without our proposed modules. Our RCA achieves an average gain of 1.1\% mIoU on RefCOCO. Next, our CPCL and TCCL achieve an average gain of 1.0\% mIoU and 0.6\% mIoU, and further boost the performance gain up to 2.7\% mIoU compared with the baseline model. Moreover, it's seen that CPCL and TCCL significantly reduce the NTA-IoU by up to 3.7\% mIoU and 4.5\% mIoU, indicating that our TLM can effectively address the `NTA' issue. These improvements enable our TALENT to set the new SOTA performance for PET-based RIS and show that our method can precisely segment the text-referred target. 

\begin{table}[h]
\centering
\renewcommand{\arraystretch}{0.85}
\small
\setlength\tabcolsep{1.5pt}
\caption{Ablation results on each proposed module to validate the performance impact using the mIoU and NTA-IoU metrics.}
\begin{tabular}{c|c|c|ccc|c|c}
\toprule[0.3mm]
\multirow{2}{*}{RCA}  & \multirow{2}{*}{CPCL} & \multirow{2}{*}{TCCL} & \multicolumn{3}{c|}{RefCOCO-mIoU(\%)}                                   & \multirow{2}{*}{AVG(\%)} & \multirow{2}{*}{NTA-IoU(\%)} \\ \cline{4-6}
                      &                      &                      & \multicolumn{1}{c|}{val}  & \multicolumn{1}{c|}{testA} & testB &                      \\ \midrule
    \ding{56}         &  \ding{56}           &      \ding{56}       & \multicolumn{1}{c|}{74.9} & \multicolumn{1}{c|}{77.1}  & 71.9  & 74.6  & 9.9              \\
    \ding{52}         &  \ding{56}           &      \ding{56}       & \multicolumn{1}{c|}{76.0} & \multicolumn{1}{c|}{77.9}  & 73.3  & 75.7 & 8.2               \\
    \ding{52}         &  \ding{52}           &      \ding{56}       & \multicolumn{1}{c|}{77.2} & \multicolumn{1}{c|}{78.7}  & 74.1  & 76.7 &  4.5             \\
    \ding{52}         &  \ding{56}           &      \ding{52}       & \multicolumn{1}{c|}{76.8} & \multicolumn{1}{c|}{78.4}  & 73.8  & 76.3 & 3.7               \\
    \ding{52}         &  \ding{52}           &      \ding{52}       & \multicolumn{1}{c|}{\textbf{77.8}} & \multicolumn{1}{c|}{\textbf{79.4}}  & \textbf{74.8}  & \textbf{77.3} &   \textbf{2.1}             \\ 
\bottomrule[0.3mm]
\end{tabular}
\label{tab4}
\end{table}
\noindent
\textbf{Effect on feature aggregation architectures for PET.}
In Tab.~\ref{tab5}, we compare different feature aggregation architectures of Eq.~(\ref{eq:4}) and the verification of the rescaling diagonal matrix ($\varphi$). 
It's seen that setting \#6 and \#4 both surpass setting \#5 and \#3 by 0.7\% mIoU on testB set, indicating that the learnable diagonal matrix contributes to the robust tuning on visual-text interaction. It's also shown that directly adding the residual feature performs better than the concatenation operation. One possible reason is that these two features are mixed in the concatenated channel dimension, which implicitly hinders the tuning process. Moreover, Settings \#1 and \#2 both suffer from a significant performance drop, which indicates the importance of the propagated vision feature $f_v^i$. These results show the effectiveness of both feature aggregation strategy and rescaling diagonal matrix.

\begin{table}[!htbp]
\centering
\setlength\tabcolsep{7.0pt}
\caption{Ablation results on different feature aggregation architectures of Eq.~\ref{eq:4}.
$cat(\cdot)$ means to concatenate the two input features and then use an MLP for channel reduction.}
\small
\renewcommand{\arraystretch}{0.9}
\begin{tabular}{cl|ccc}
\toprule[0.3mm]
\multirow{2}{*}{Settings} & \multirow{2}{*}{Architecture}  & \multicolumn{3}{c}{RefCOCO-mIoU(\%)} \\ \cline{3-5}   &                          & val    & testA    & testB   \\ 
\midrule
\#1           & $\nu_{i+1}(f_{vt}^{i})$                          &  71.8  &  76.0    &  70.1  \\
\#2           & $\nu_{i+1}(f_{vt}^{i})*\varphi$                  &  71.1  &  74.5    &  68.9  \\
\#3           & $cat(\nu_{i+1}(f_{v}^{i}), f_{vt}^{i})$          &  76.1  &  77.9    &  73.2  \\
\#4           & $cat(\nu_{i+1}(f_{v}^{i}), f_{vt}^{i}*\varphi)$  &  76.6  &  78.8    &  73.9  \\
\#5           & $\nu_{i+1}(f_{v}^{i}) + f_{vt}^{i}$              &  77.1  &  78.7    &  74.1  \\
\#6           & $\nu_{i+1}(f_{v}^{i}) + f_{vt}^{i}*\varphi$      &  \textbf{77.8}  &  \textbf{79.4}    &  \textbf{74.8}  \\
\bottomrule[0.3mm]
\end{tabular}
\label{tab5}
\end{table}

\noindent
\textbf{Effect on sensitivity of the loss scaling coefficients.}
Tab.~\ref{tab6} evaluates the effects of scaling coefficient $\lambda_{cpcl}$ and $\lambda_{tccl}$ on the RefCOCO dataset. Specifically, when $\lambda_{cpcl}$ and $\lambda_{tccl}$ are both set to 0.1, TALENT achieves the best performance. It indicates that our progressive context learning serves as an auxiliary loss that will not impact the main loss. These two auxiliary losses improve the segmentation performance and mitigate the `NTA' issue without extra trainable parameters. More results are summarized in the appendix.

\begin{table}[!htbp]
\centering
\setlength\tabcolsep{5.0pt}
\small
\renewcommand{\arraystretch}{0.85}
\caption{Ablation results on each loss scaling coefficient.}
\begin{tabular}{l|cccccc|cc}
\toprule[0.3mm]
$\lambda_{cpcl}$   & 0.3  & 0.2  & 0.2  & 0.1  & 0.1  & 0.1 & 0.1  & 0.0  \\
$\lambda_{tccl}$   & 0.1  & 0.2  & 0.1  & 0.3  & 0.2  & 0.1 & 0.0  & 0.1 \\ \midrule
val   & 76.9 & 77.2 & 77.5 & 77.4 & 77.6 & \textbf{77.8} & 77.2  & 76.8  \\
testA & 78.5 & 78.6 & 79.1 & 78.9 & 79.1 & \textbf{79.4} & 78.7  & 78.4  \\
testB & 74.3 & 74.3 & 74.4 & 74.2 & 74.6 & \textbf{74.8} & 74.1  & 73.8  \\ 
\bottomrule[0.3mm]
\end{tabular}
\label{tab6}
\end{table}

\section{Conclusion}
In this paper, we identify `NTA' issue in PET-based RIS methods and introduce a new metric, NTA-IoU, to quantify its impact. To address `NTA', we propose TALENT, an efficient tuning framework. First, TALENT introduces a Rectified Cost Aggregator (RCA) for visual-text interaction. Next, TALENT adopts a Target-aware Learning Mechanism (TLM), including Contextual Pairwise Consistency Learning (CPCL) and Target Centric Contrastive Learning (TCCL). CPCL optimizes the visual feature representation guided by a text-augmented affinity map, and TCCL further improves its discriminative ability between the distinct target instance and other irrelevant ones. Extensive experiments indicate that TALENT achieves significant gains over existing methods and effectively mitigates the `NTA' issue.

\section{Acknowledgment}
This work was supported by the National Natural Science Foundation of China (No. 62301451, 62301613, 62471405, 62331003), Basic Research Program of Jiangsu (BK20241814), Suzhou Basic Research Program (SYG202316), XJTLU REF-22-01-010 and XJTLU RDF-22-02-066.

\clearpage
\section{Supplementary}
In this supplementary material, we provide additional ablation experiments of the target-aware learning mechanism in Sec.~\ref{sec:1} with further discussions, and provide more feature activation and segmentation results in Sec.~\ref{sec:2}.

\section{More ablation experiments}
\label{sec:1}
\subsection{Analysis about target-aware learning mechanism upon visual-text interaction.}

To further analyze the effectiveness of our Target-aware Learning Mechanism (TLM), including Contextual Pairwise Consistency Learning (CPCL) and Target-Centric Contrastive Learning (TCCL), we conduct experiments by applying them to another visual-text interaction approach, cross-attention. For a fair comparison, we keep all other components consistent with our Rectified Cost Aggregator (RCA). Note that the cross attention is a standard design, where the image is denoted as the query, while the text is denoted as the key and the value. 

As summarized in Table~\ref{supp:tab1}, we first discard the visual-text interaction mechanisms between visual and text backbones and take it as the baseline setting. It's observed that this setting can't achieve promising performance. Then, we compare the performance of different visual-text interaction mechanisms. Specifically, when only evaluating RCA against cross-attention in isolation, \ie, without TLM, RCA achieves 1.4\% mIoU gains over the baseline and 0.9\% mIoU gains over the cross-attention on the RefCOCO testB set. When both CPCL and TCCL are applied, RCA further achieves 2.9\% mIoU gains over the baseline and 0.7\% mIoU gains over the cross-attention mechanism, with fewer parameter numbers as shown in Tab. 3 of the main paper.

In summary, RCA represents the first attempt to leverage cost-volume analysis for visual–text interaction modeling in PET-based RIS. These results demonstrate that our RCA serves as a more effective method for visual-text interaction compared to the cross-attention mechanism. 

\noindent
\textbf{Discussion about generalization of TLM}. Recall that TLM is designed to refine the multimodal features produced from visual–text interaction modules for mitigating the `NTA' issue. Therefore, TLM is expected to possess a strong generalization ability to handle multimodal features produced by different visual–text interaction mechanisms. As shown in Tab.~\ref {supp:tab1}, TLM also functions effectively when combined with cross-attention instead of RCA. Specifically, when only using cross-attention in isolation, \ie, without TLM, it only achieves an average of 0.4\% mIoU gains over the baseline on RefCOCO. When both CPCL and TCCL are applied, it further achieves 2.2\% mIoU gains over the baseline model. These results underscore the generalization ability of our TLM, which also works on cross-attention.

\begin{table}[!ht]
\centering
\small
\setlength\tabcolsep{4.5pt}
\caption{Ablation experiments on applying CPCL and TCCL to different visual-text interaction mechanisms.}
\begin{tabular}{c|cc|ccc|c}
\toprule[0.3mm]
\multirow{2}{*}{V-L interaction} & \multicolumn{2}{c|}{TLM} & \multicolumn{3}{c|}{RefCOCO} & \multirow{2}{*}{Avg} \\ \cline{2-6}
                    &    CPCL    &    TCCL   & val    & testA    & testB     \\ \hline
\ding{56}     &  \ding{56}    &  \ding{56}   &  74.9  &  77.1    & 71.9   & 74.6   \\
RCA     &  \ding{56}    &  \ding{56}   &  76.0  &  77.9    & 73.3   & 75.7   \\
RCA     &  \ding{52}    &  \ding{52}   &  \textbf{77.8}  &  \textbf{79.4}    & \textbf{74.8}  & \textbf{77.3}  \\ \hline
cross-attention     &  \ding{56}    &  \ding{56}   &  75.4  &  77.3    & 72.4  & 75.0  \\ 
cross-attention     &  \ding{52}    &  \ding{52}   &  77.3  &  78.9    & 74.1  & 76.8  \\
\bottomrule[0.3mm]
\end{tabular}
\label{supp:tab1}
\end{table}

\subsection{Effect on sensitivity of loss scaling coefficient}
\begin{figure}[!ht]
\centering
\includegraphics[width=0.4\textwidth]{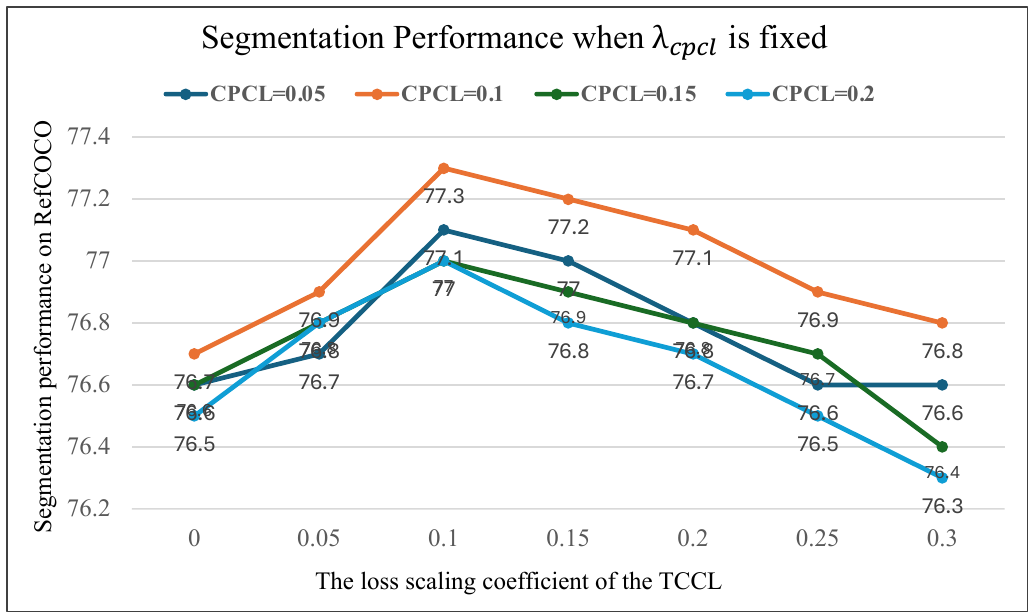}
\caption{Ablation results of the evaluation of the loss scaling coefficient for our CPCL and TCCL. Each colored line corresponds to a specific CPCL coefficient.}
\label{supp:fig1}
\end{figure}

To provide a more comprehensive analysis of the two learning objectives in TLM, we try to fix one coefficient of the two objectives and verify how they influence the segmentation performance. Specifically, we fix the coefficient of CPCL to several representative values and vary the coefficient of TCCL accordingly. The performance curves are plotted in Fig.~\ref{supp:fig1}, where each colored line corresponds to a specific CPCL coefficient. It is observed that when $\lambda_{\text{cpcl}}$ is fixed at 0.1, TALENT consistently achieves the best performance on the RefCOCO dataset, regardless of the value of the TCCL coefficient. Moreover, among different values of $\lambda_{\text{cpcl}}$, TALENT tends to achieve optimal performance when $\lambda_{\text{tccl}}$ is set to 0.1. These results indicate that setting both $\lambda_{\text{cpcl}}$ and $\lambda_{\text{tccl}}$ to 0.1 yields the best performance.

\begin{figure*}[t]
\centering
\includegraphics[width=0.9\textwidth]{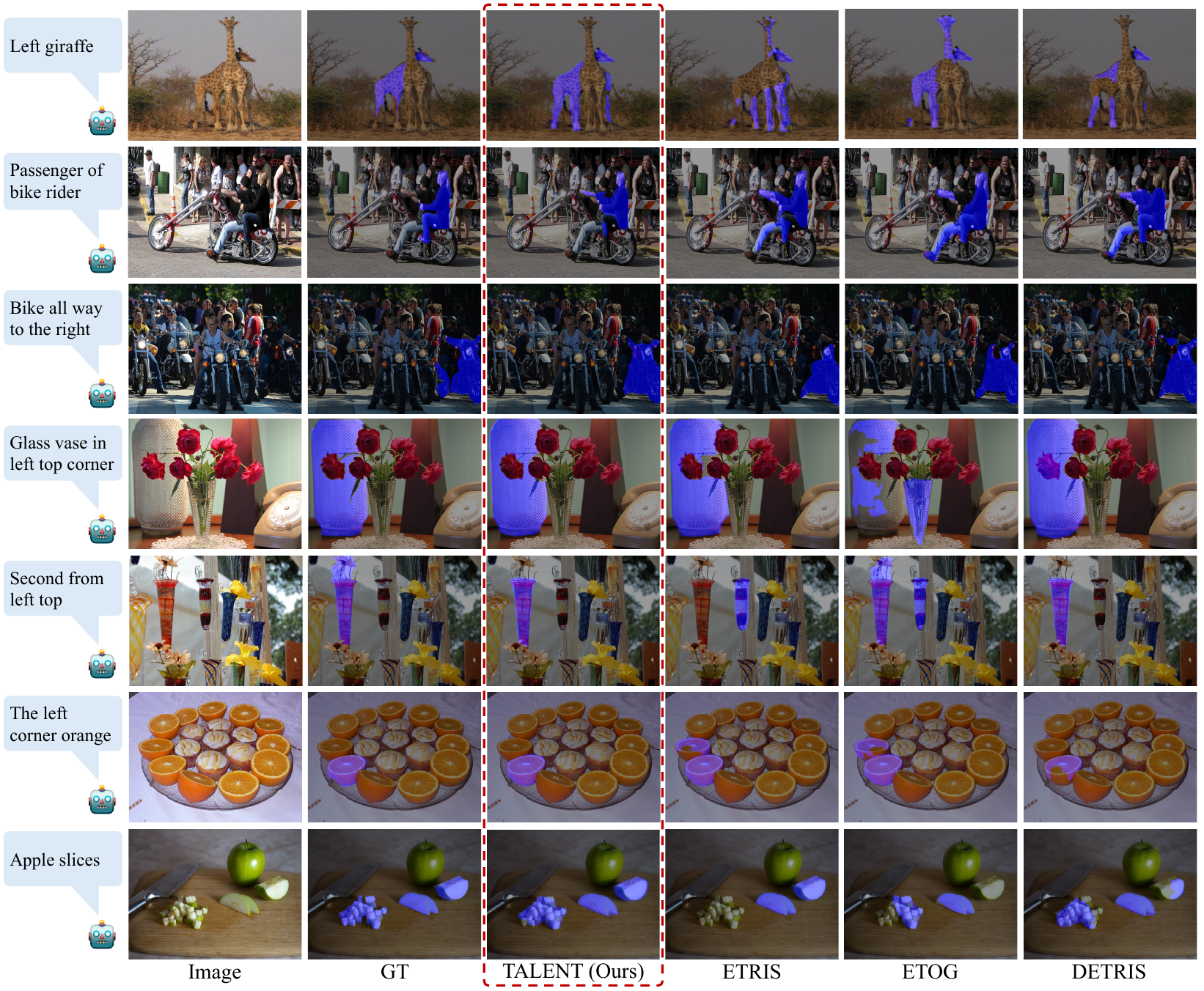}
\caption{Visualization of segmentation results. We compare our TALENT with existing PET-based methods. It's observed that TALENT can generate more precise segmentation maps.}
\label{supp:fig3}
\end{figure*}

\section{More visualization results}
\label{sec:2}
\subsection{Segmentation visualization}
We compare more visualization results of the segmentation maps with PET-based methods~\cite{huang2025densely,xu2023bridging,yu2024etog}, which are illustrated in Fig.~\ref{supp:fig3}. It's evident that prior approaches have difficulty accurately localizing the target instance described by the text expression. Instead, these methods frequently misidentify other salient and similar objects, like `apple slices' and `left corner orange' in the last and the penultimate row of Fig.~\ref{supp:fig3}. In contrast, our TALENT can accurately identify and segment the distinct text-referred target instance. These results demonstrate that our proposed target-aware learning mechanism can effectively enhance the feature discrimination and reduce `NTA' impacts.

\begin{figure}[!ht]
\centering
\includegraphics[width=0.4\textwidth]{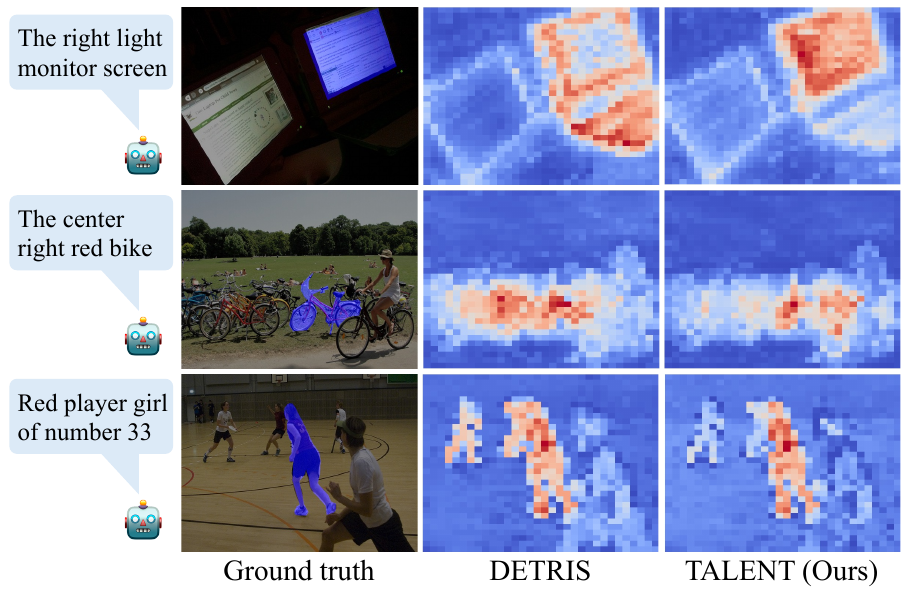}
\caption{Visualization of feature activation maps. We compare TALENT with the SOTA PET-based method DETRIS~\cite{huang2025densely}.}
\label{supp:fig2}
\end{figure}

\subsection{Feature activation visualization}
Fig.~\ref{supp:fig2} compares the feature activation among different PET-based methods to evaluate the effectiveness of mitigating the `NTA' issue. It's observed that our TALENT can accurately activate the distinct text-referred visual region, whereas the previous SOTA method, DETRIS~\cite{huang2025densely}, often activates salient yet unrelated objects. Specifically, given the text expression `the center right red bike' in the second row, DETRIS~\cite{huang2025densely} often tends to activate other similar objects, \eg, other red bikes. In the third row, DETRIS~\cite{huang2025densely} tends to activate other player girls in red when given the text expression `red player girl of number 33'. In contrast, our TALENT selectively activates the specific target that consistently aligns with the GT segmentation mask, which underscores the importance of mitigating `NTA'.



\begin{thebibliography}{62}
\providecommand{\natexlab}[1]{#1}
\providecommand{\url}[1]{\texttt{#1}}
\expandafter\ifx\csname urlstyle\endcsname\relax
  \providecommand{\doi}[1]{doi: #1}\else
  \providecommand{\doi}{doi: \begingroup \urlstyle{rm}\Url}\fi

\bibitem[Chen et~al.(2022)Chen, Hu, Ji, Bai, and Zuo]{chen2022position}
Bo Chen, Zhiwei Hu, Zhilong Ji, Jinfeng Bai, and Wangmeng Zuo.
\newblock Position-aware contrastive alignment for referring image segmentation.
\newblock \emph{arXiv preprint arXiv:2212.13419}, 2022.

\bibitem[Cheng et~al.(2025)Cheng, Liu, He, Ourselin, Tan, and Luo]{cheng2025weakmcn}
Silin Cheng, Yang Liu, Xinwei He, Sebastien Ourselin, Lei Tan, and Gen Luo.
\newblock Weakmcn: Multi-task collaborative network for weakly supervised referring expression comprehension and segmentation.
\newblock In \emph{CVPR}, pages 9175--9185, 2025.

\bibitem[Chng et~al.(2024)Chng, Zheng, Han, Qiu, and Huang]{chng2023mask}
Yong~Xien Chng, Henry Zheng, Yizeng Han, Xuchong Qiu, and Gao Huang.
\newblock Mask grounding for referring image segmentation.
\newblock In \emph{CVPR}, pages 26573--26583, 2024.

\bibitem[Cho et~al.(2024)Cho, Shin, Hong, Arnab, Seo, and Kim]{cho2024cat}
Seokju Cho, Heeseong Shin, Sunghwan Hong, Anurag Arnab, Paul~Hongsuck Seo, and Seungryong Kim.
\newblock Cat-seg: Cost aggregation for open-vocabulary semantic segmentation.
\newblock In \emph{CVPR}, pages 4113--4123, 2024.

\bibitem[Dai et~al.(2025)Dai, Li, Zhuang, Zhang, and Yang]{dai2025multi}
Ming Dai, Jian Li, Jiedong Zhuang, Xian Zhang, and Wankou Yang.
\newblock Multi-task visual grounding with coarse-to-fine consistency constraints.
\newblock In \emph{AAAI}, pages 2618--2626, 2025.

\bibitem[Ding et~al.(2021)Ding, Liu, Wang, and Jiang]{vlt}
Henghui Ding, Chang Liu, Suchen Wang, and Xudong Jiang.
\newblock Vision-language transformer and query generation for referring segmentation.
\newblock In \emph{ICCV}, pages 16321--16330, 2021.

\bibitem[Fang et~al.(2023)Fang, Wang, Xie, Sun, Wu, Wang, Huang, Wang, and Cao]{fang2023eva}
Yuxin Fang, Wen Wang, Binhui Xie, Quan Sun, Ledell Wu, Xinggang Wang, Tiejun Huang, Xinlong Wang, and Yue Cao.
\newblock Eva: Exploring the limits of masked visual representation learning at scale.
\newblock In \emph{CVPR}, pages 19358--19369, 2023.

\bibitem[Gao et~al.(2024)Gao, Geng, Zhang, Ma, Fang, Zhang, Li, and Qiao]{gao2024clipadapter}
Peng Gao, Shijie Geng, Renrui Zhang, Teli Ma, Rongyao Fang, Yongfeng Zhang, Hongsheng Li, and Yu Qiao.
\newblock Clip-adapter: Better vision-language models with feature adapters.
\newblock \emph{IJCV}, 132\penalty0 (2):\penalty0 581--595, 2024.

\bibitem[He et~al.(2016)He, Zhang, Ren, and Sun]{resnet}
Kaiming He, Xiangyu Zhang, Shaoqing Ren, and Jian Sun.
\newblock Deep residual learning for image recognition.
\newblock In \emph{CVPR}, pages 770--778, 2016.

\bibitem[He et~al.(2022)He, Chen, Xie, Li, Doll{\'a}r, and Girshick]{he2022masked}
Kaiming He, Xinlei Chen, Saining Xie, Yanghao Li, Piotr Doll{\'a}r, and Ross Girshick.
\newblock Masked autoencoders are scalable vision learners.
\newblock In \emph{CVPR}, pages 16000--16009, 2022.

\bibitem[Hu et~al.(2022)Hu, Shen, Wallis, Allen-Zhu, Li, Wang, Wang, Chen, et~al.]{hu2021lora}
Edward~J Hu, Yelong Shen, Phillip Wallis, Zeyuan Allen-Zhu, Yuanzhi Li, Shean Wang, Lu Wang, Weizhu Chen, et~al.
\newblock Lora: Low-rank adaptation of large language models.
\newblock In \emph{ICLR}, pages 1--8, 2022.

\bibitem[Hu et~al.(2016)Hu, Rohrbach, and Darrell]{RIS1}
Ronghang Hu, Marcus Rohrbach, and Trevor Darrell.
\newblock Segmentation from natural language expressions.
\newblock In \emph{ECCV}, pages 108--124, 2016.

\bibitem[Hu et~al.(2023)Hu, Wang, Shao, Xie, Li, Han, and Luo]{hu2023onetoonerethinkingreferringimage}
Yutao Hu, Qixiong Wang, Wenqi Shao, Enze Xie, Zhenguo Li, Jungong Han, and Ping Luo.
\newblock Beyond one-to-one: Rethinking the referring image segmentation.
\newblock In \emph{ICCV}, pages 4067--4077, 2023.

\bibitem[Hu et~al.(2020)Hu, Feng, Sun, Zhang, and Lu]{brinet}
Zhiwei Hu, Guang Feng, Jiayu Sun, Lihe Zhang, and Huchuan Lu.
\newblock Bi-directional relationship inferring network for referring image segmentation.
\newblock In \emph{CVPR}, pages 4424--4433, 2020.

\bibitem[Huang et~al.(2025{\natexlab{a}})Huang, Xu, Liu, Liu, Han, Yuan, and Li]{huang2025densely}
Jiaqi Huang, Zunnan Xu, Ting Liu, Yong Liu, Haonan Han, Kehong Yuan, and Xiu Li.
\newblock Densely connected parameter-efficient tuning for referring image segmentation.
\newblock In \emph{AAAI}, pages 3653--3661, 2025{\natexlab{a}}.

\bibitem[Huang et~al.(2025{\natexlab{b}})Huang, Fu, Liu, Jiang, Yu, and Song]{Huang_Fu_Liu_Jiang_Yu_Song_2025}
Qihan Huang, Siming Fu, Jinlong Liu, Hao Jiang, Yipeng Yu, and Jie Song.
\newblock Resolving multi-condition confusion for finetuning-free personalized image generation.
\newblock In \emph{AAAI}, pages 3707--3714, 2025{\natexlab{b}}.

\bibitem[Huang et~al.(2020)Huang, Hui, Liu, Li, Wei, Han, Liu, and Li]{huang2020referring}
Shaofei Huang, Tianrui Hui, Si Liu, Guanbin Li, Yunchao Wei, Jizhong Han, Luoqi Liu, and Bo Li.
\newblock Referring image segmentation via cross-modal progressive comprehension.
\newblock In \emph{CVPR}, pages 10488--10497, 2020.

\bibitem[Jie et~al.(2024)Jie, Deng, Chen, and Jin]{jie2022convolutional}
Shibo Jie, Zhi-Hong Deng, Shixuan Chen, and Zhijuan Jin.
\newblock Convolutional bypasses are better vision transformer adapters.
\newblock In \emph{ECAI}, pages 202--209, 2024.

\bibitem[Jin et~al.(2023)Jin, Liu, Yao, Lin, and Zhao]{jin2023kernel}
Shuo Jin, Meiqin Liu, Chao Yao, Chunyu Lin, and Yao Zhao.
\newblock Kernel dimension matters: To activate available kernels for real-time video super-resolution.
\newblock In \emph{ACMMM}, pages 8617--8625, 2023.

\bibitem[Jin et~al.(2025)Jin, Yu, Zhang, Sun, Dong, and Xiao]{jin2025feature}
Shuo Jin, Siyue Yu, Bingfeng Zhang, Mingjie Sun, Yi Dong, and Jimin Xiao.
\newblock Feature purification matters: Suppressing outlier propagation for training-free open-vocabulary semantic segmentation.
\newblock In \emph{ICCV}, pages 20291--20300, 2025.

\bibitem[Karimi~Mahabadi et~al.(2021)Karimi~Mahabadi, Henderson, and Ruder]{karimi2021compacter}
Rabeeh Karimi~Mahabadi, James Henderson, and Sebastian Ruder.
\newblock Compacter: Efficient low-rank hypercomplex adapter layers.
\newblock In \emph{NeurIPS}, pages 1022--1035, 2021.

\bibitem[Kazemzadeh et~al.(2014)Kazemzadeh, Ordonez, Matten, and Berg]{refcoco}
Sahar Kazemzadeh, Vicente Ordonez, Mark Matten, and Tamara Berg.
\newblock Referitgame: Referring to objects in photographs of natural scenes.
\newblock In \emph{EMNLP}, pages 787--798, 2014.

\bibitem[Kim et~al.(2024)Kim, Kang, Kim, Park, and Kwak]{kim2024extendingclipsimagetextalignment}
Seoyeon Kim, Minguk Kang, Dongwon Kim, Jaesik Park, and Suha Kwak.
\newblock Extending clip's image-text alignment to referring image segmentation.
\newblock In \emph{NAACL}, pages 4611--4628, 2024.

\bibitem[Kirillov et~al.(2023)Kirillov, Mintun, Ravi, Mao, Rolland, Gustafson, Xiao, Whitehead, Berg, Lo, et~al.]{sam}
Alexander Kirillov, Eric Mintun, Nikhila Ravi, Hanzi Mao, Chloe Rolland, Laura Gustafson, Tete Xiao, Spencer Whitehead, Alexander~C Berg, Wan-Yen Lo, et~al.
\newblock Segment anything.
\newblock In \emph{ICCV}, pages 4015--4026, 2023.

\bibitem[Lai et~al.(2024)Lai, Tian, Chen, Li, Yuan, Liu, and Jia]{lai2024lisa}
Xin Lai, Zhuotao Tian, Yukang Chen, Yanwei Li, Yuhui Yuan, Shu Liu, and Jiaya Jia.
\newblock Lisa: Reasoning segmentation via large language model.
\newblock In \emph{CVPR}, pages 9579--9589, 2024.

\bibitem[Li et~al.(2023)Li, Sun, Xiao, Lim, and Zhao]{li2023fully}
Hui Li, Mingjie Sun, Jimin Xiao, Eng~Gee Lim, and Yao Zhao.
\newblock Fully and weakly supervised referring expression segmentation with end-to-end learning.
\newblock \emph{IEEE T-CSVT}, 33\penalty0 (10):\penalty0 5999--6012, 2023.

\bibitem[Lin et~al.(2014)Lin, Maire, Belongie, Hays, Perona, Ramanan, Doll{\'a}r, and Zitnick]{lin2014microsoft}
Tsung-Yi Lin, Michael Maire, Serge Belongie, James Hays, Pietro Perona, Deva Ramanan, Piotr Doll{\'a}r, and C~Lawrence Zitnick.
\newblock Microsoft coco: Common objects in context.
\newblock In \emph{ECCV}, pages 740--755, 2014.

\bibitem[Liu et~al.(2023{\natexlab{a}})Liu, Ding, and Jiang]{gres}
Chang Liu, Henghui Ding, and Xudong Jiang.
\newblock Gres: Generalized referring expression segmentation.
\newblock In \emph{CVPR}, pages 23592--23601, 2023{\natexlab{a}}.

\bibitem[Liu et~al.(2022)Liu, Jin, Yao, Lin, and Zhao]{liu2022temporal}
Meiqin Liu, Shuo Jin, Chao Yao, Chunyu Lin, and Yao Zhao.
\newblock Temporal consistency learning of inter-frames for video super-resolution.
\newblock \emph{IEEE T-CSVT}, 33\penalty0 (4):\penalty0 1507--1520, 2022.

\bibitem[Liu et~al.(2021{\natexlab{a}})Liu, Hui, Huang, Wei, Li, and Li]{cmpc+}
Si Liu, Tianrui Hui, Shaofei Huang, Yunchao Wei, Bo Li, and Guanbin Li.
\newblock Cross-modal progressive comprehension for referring segmentation.
\newblock \emph{IEEE T-PAMI}, 44\penalty0 (9):\penalty0 4761--4775, 2021{\natexlab{a}}.

\bibitem[Liu et~al.(2023{\natexlab{b}})Liu, Zhang, Qiu, Xie, Zhang, and Yao]{liu2023caris}
Sun-Ao Liu, Yiheng Zhang, Zhaofan Qiu, Hongtao Xie, Yongdong Zhang, and Ting Yao.
\newblock Caris: Context-aware referring image segmentation.
\newblock In \emph{ACMMM}, pages 779--788, 2023{\natexlab{b}}.

\bibitem[Liu et~al.(2021{\natexlab{b}})Liu, Lin, Cao, Hu, Wei, Zhang, Lin, and Guo]{swin}
Ze Liu, Yutong Lin, Yue Cao, Han Hu, Yixuan Wei, Zheng Zhang, Stephen Lin, and Baining Guo.
\newblock Swin transformer: Hierarchical vision transformer using shifted windows.
\newblock In \emph{ICCV}, pages 10012--10022, 2021{\natexlab{b}}.

\bibitem[Long et~al.(2015)Long, Shelhamer, and Darrell]{fcn}
Jonathan Long, Evan Shelhamer, and Trevor Darrell.
\newblock Fully convolutional networks for semantic segmentation.
\newblock In \emph{CVPR}, pages 3431--3440, 2015.

\bibitem[Luo et~al.(2020)Luo, Zhou, Sun, Cao, Wu, Deng, and Ji]{mcn}
Gen Luo, Yiyi Zhou, Xiaoshuai Sun, Liujuan Cao, Chenglin Wu, Cheng Deng, and Rongrong Ji.
\newblock Multi-task collaborative network for joint referring expression comprehension and segmentation.
\newblock In \emph{CVPR}, pages 10034--10043, 2020.

\bibitem[Luo et~al.(2025)Luo, Wu, Cheng, Liu, Xiao, Wang, Zhang, and Yang]{luo2024cohd}
Zhuoyan Luo, Yinghao Wu, Tianheng Cheng, Yong Liu, Yicheng Xiao, Hongfa Wang, Xiao-Ping Zhang, and Yujiu Yang.
\newblock Cohd: A counting-aware hierarchical decoding framework for generalized referring expression segmentation.
\newblock In \emph{ICCV}, pages 1--8, 2025.

\bibitem[Nagaraja et~al.(2016)Nagaraja, Morariu, and Davis]{grefcoco}
Varun~K Nagaraja, Vlad~I Morariu, and Larry~S Davis.
\newblock Modeling context between objects for referring expression understanding.
\newblock In \emph{ECCV}, pages 792--807, 2016.

\bibitem[Oquab et~al.(2023)Oquab, Darcet, Moutakanni, Vo, Szafraniec, Khalidov, Fernandez, HAZIZA, Massa, El-Nouby, et~al.]{oquab2023dinov2}
Maxime Oquab, Timoth{\'e}e Darcet, Th{\'e}o Moutakanni, Huy~V Vo, Marc Szafraniec, Vasil Khalidov, Pierre Fernandez, Daniel HAZIZA, Francisco Massa, Alaaeldin El-Nouby, et~al.
\newblock Dinov2: Learning robust visual features without supervision.
\newblock \emph{TMLR}, 2023.

\bibitem[Pan et~al.(2024)Pan, Sun, Wang, Zhang, and Zhang]{pan2024rethinking}
Yuwen Pan, Rui Sun, Yuan Wang, Tianzhu Zhang, and Yongdong Zhang.
\newblock Rethinking the implicit optimization paradigm with dual alignments for referring remote sensing image segmentation.
\newblock In \emph{ACMMM}, pages 2031--2040, 2024.

\bibitem[Peng et~al.(2025)Peng, Xu, Zeng, Huang, Wang, and Shen]{peng2025parameter}
Zelin Peng, Zhengqin Xu, Zhilin Zeng, Yu Huang, Yaoming Wang, and Wei Shen.
\newblock Parameter-efficient fine-tuning in hyperspherical space for open-vocabulary semantic segmentation.
\newblock In \emph{CVPR}, pages 15009--15020, 2025.

\bibitem[Qiu et~al.(2025)Qiu, Wang, Zhang, and Xiao]{qiu2025bias}
Xianglin Qiu, Xiaoyang Wang, Zhen Zhang, and Jimin Xiao.
\newblock Bias-resilient weakly supervised semantic segmentation using normalizing flows.
\newblock In \emph{CVPR}, pages 21321--21330, 2025.

\bibitem[Radford et~al.(2021{\natexlab{a}})Radford, Kim, Hallacy, Ramesh, Goh, Agarwal, Sastry, Askell, Mishkin, Clark, et~al.]{clip}
Alec Radford, Jong~Wook Kim, Chris Hallacy, Aditya Ramesh, Gabriel Goh, Sandhini Agarwal, Girish Sastry, Amanda Askell, Pamela Mishkin, Jack Clark, et~al.
\newblock Learning transferable visual models from natural language supervision.
\newblock In \emph{ICML}, pages 8748--8763, 2021{\natexlab{a}}.

\bibitem[Radford et~al.(2021{\natexlab{b}})Radford, Kim, Hallacy, Ramesh, Goh, Agarwal, Sastry, Askell, Mishkin, Clark, et~al.]{radford2021learning}
Alec Radford, Jong~Wook Kim, Chris Hallacy, Aditya Ramesh, Gabriel Goh, Sandhini Agarwal, Girish Sastry, Amanda Askell, Pamela Mishkin, Jack Clark, et~al.
\newblock Learning transferable visual models from natural language supervision.
\newblock In \emph{ICML}, pages 8748--8763, 2021{\natexlab{b}}.

\bibitem[Shah et~al.(2024)Shah, VS, and Patel]{shah2024lqmformer}
Nisarg~A Shah, Vibashan VS, and Vishal~M Patel.
\newblock Lqmformer: Language-aware query mask transformer for referring image segmentation.
\newblock In \emph{CVPR}, pages 12903--12913, 2024.

\bibitem[Shang et~al.(2024)Shang, Song, Qiu, Wang, Meng, and Li]{shang2024prompt}
Chao Shang, Zichen Song, Heqian Qiu, Lanxiao Wang, Fanman Meng, and Hongliang Li.
\newblock Prompt-driven referring image segmentation with instance contrasting.
\newblock In \emph{CVPR}, pages 4124--4134, 2024.

\bibitem[Sun et~al.(2021{\natexlab{a}})Sun, Xiao, and Lim]{sun2021iterative}
Mingjie Sun, Jimin Xiao, and Eng~Gee Lim.
\newblock Iterative shrinking for referring expression grounding using deep reinforcement learning.
\newblock In \emph{CVPR}, pages 14060--14069, 2021{\natexlab{a}}.

\bibitem[Sun et~al.(2021{\natexlab{b}})Sun, Xiao, Lim, Liu, and Goulermas]{sun2021discriminative}
Mingjie Sun, Jimin Xiao, Eng~Gee Lim, Si Liu, and John~Y Goulermas.
\newblock Discriminative triad matching and reconstruction for weakly referring expression grounding.
\newblock \emph{IEEE T-PAMI}, 43\penalty0 (11):\penalty0 4189--4195, 2021{\natexlab{b}}.

\bibitem[Tang et~al.(2025{\natexlab{a}})Tang, Huang, Liu, Sun, Yang, and Lim]{tang2025intervening}
Feilong Tang, Zile Huang, Chengzhi Liu, Qiang Sun, Harry Yang, and Ser-Nam Lim.
\newblock Intervening anchor token: Decoding strategy in alleviating hallucinations for mllms.
\newblock In \emph{ICLR}, pages 1--8, 2025{\natexlab{a}}.

\bibitem[Tang et~al.(2025{\natexlab{b}})Tang, Liu, Xu, Hu, Huang, Xue, Chen, Peng, Yang, Zhou, et~al.]{tang2025seeing}
Feilong Tang, Chengzhi Liu, Zhongxing Xu, Ming Hu, Zile Huang, Haochen Xue, Ziyang Chen, Zelin Peng, Zhiwei Yang, Sijin Zhou, et~al.
\newblock Seeing far and clearly: Mitigating hallucinations in mllms with attention causal decoding.
\newblock In \emph{CVPR}, pages 26147--26159, 2025{\natexlab{b}}.

\bibitem[Tang et~al.(2023)Tang, Zheng, Shi, and Yang]{tang2023contrastive}
Jiajin Tang, Ge Zheng, Cheng Shi, and Sibei Yang.
\newblock Contrastive grouping with transformer for referring image segmentation.
\newblock In \emph{CVPR}, pages 23570--23580, 2023.

\bibitem[Touvron et~al.(2021)Touvron, Cord, Sablayrolles, Synnaeve, and J{\'e}gou]{touvron2021going}
Hugo Touvron, Matthieu Cord, Alexandre Sablayrolles, Gabriel Synnaeve, and Herv{\'e} J{\'e}gou.
\newblock Going deeper with image transformers.
\newblock In \emph{ICCV}, pages 32--42, 2021.

\bibitem[Wang et~al.(2023)Wang, Li, ZHANG, Shi, Li, Dai, Xiong, and Tian]{wang2023barleria}
Yaoming Wang, Jin Li, XIAOPENG ZHANG, Bowen Shi, Chenglin Li, Wenrui Dai, Hongkai Xiong, and Qi Tian.
\newblock Barleria: An efficient tuning framework for referring image segmentation.
\newblock In \emph{ICLR}, pages 1--8, 2023.

\bibitem[Wang et~al.(2025)Wang, Wang, Gong, and Xiao]{wang2025normal}
Yuexin Wang, Xiaolei Wang, Yizheng Gong, and Jimin Xiao.
\newblock Normal-abnormal guided generalist anomaly detection.
\newblock In \emph{NeurIPS}, pages 1--8, 2025.

\bibitem[Wang et~al.(2026)Wang, Wang, Li, Liu, Tillo, and Xiao]{wang2025FGPT}
Yuexin Wang, Xiaoyang Wang, Haocheng Li, Jiejie Liu, Tammam Tillo, and Jimin Xiao.
\newblock Fgpt: Fine-grained prompt tuning for zero-shot anomaly detection.
\newblock \emph{IEEE T-AI}, pages 1--13, 2026.

\bibitem[Wang et~al.(2022)Wang, Lu, Li, Tao, Guo, Gong, and Liu]{cris}
Zhaoqing Wang, Yu Lu, Qiang Li, Xunqiang Tao, Yandong Guo, Mingming Gong, and Tongliang Liu.
\newblock Cris: Clip-driven referring image segmentation.
\newblock In \emph{CVPR}, pages 11686--11695, 2022.

\bibitem[Xu et~al.(2023)Xu, Chen, Zhang, Song, Wan, and Li]{xu2023bridging}
Zunnan Xu, Zhihong Chen, Yong Zhang, Yibing Song, Xiang Wan, and Guanbin Li.
\newblock Bridging vision and language encoders: Parameter-efficient tuning for referring image segmentation.
\newblock In \emph{ICCV}, pages 17503--17512, 2023.

\bibitem[Xue et~al.(2025)Xue, Tang, Hu, Liu, Huang, Li, Liu, Xu, Zhang, Feng, et~al.]{xue2025mmrc}
Haochen Xue, Feilong Tang, Ming Hu, Yexin Liu, Qidong Huang, Yulong Li, Chengzhi Liu, Zhongxing Xu, Chong Zhang, Chun-Mei Feng, et~al.
\newblock Mmrc: A large-scale benchmark for understanding multimodal large language model in real-world conversation.
\newblock \emph{arXiv preprint arXiv:2502.11903}, 2025.

\bibitem[Yang et~al.(2025{\natexlab{a}})Yang, Tang, Zhao, An, Hu, Li, Zhuang, Wang, Lu, Zhang, et~al.]{yang2025streamagent}
Haolin Yang, Feilong Tang, Linxiao Zhao, Xiang An, Ming Hu, Huifa Li, Xinlin Zhuang, Boqian Wang, Yifan Lu, Xiaofeng Zhang, et~al.
\newblock Streamagent: Towards anticipatory agents for streaming video understanding.
\newblock \emph{arXiv preprint arXiv:2508.01875}, 2025{\natexlab{a}}.

\bibitem[Yang et~al.(2024)Yang, Ma, Yao, Zhong, Zhang, and Wang]{yang2024remamber}
Yuhuan Yang, Chaofan Ma, Jiangchao Yao, Zhun Zhong, Ya Zhang, and Yanfeng Wang.
\newblock Remamber: Referring image segmentation with mamba twister.
\newblock In \emph{ECCV}, pages 108--126, 2024.

\bibitem[Yang et~al.(2022)Yang, Wang, Tang, Chen, Zhao, and Torr]{lavt}
Zhao Yang, Jiaqi Wang, Yansong Tang, Kai Chen, Hengshuang Zhao, and Philip~HS Torr.
\newblock Lavt: Language-aware vision transformer for referring image segmentation.
\newblock In \emph{CVPR}, pages 18155--18165, 2022.

\bibitem[Yang et~al.(2025{\natexlab{b}})Yang, Zhao, Wang, Zhang, and Xiao]{yang2025ffr}
Ziqian Yang, Xinqiao Zhao, Xiaolei Wang, Quan Zhang, and Jimin Xiao.
\newblock Ffr: Frequency feature rectification for weakly supervised semantic segmentation.
\newblock In \emph{CVPR}, pages 30261--30270, 2025{\natexlab{b}}.

\bibitem[Yu et~al.(2024)Yu, Li, Rezazadeh, Yang, and Choi]{yu2024etog}
Houjian Yu, Mingen Li, Alireza Rezazadeh, Yang Yang, and Changhyun Choi.
\newblock A parameter-efficient tuning framework for language-guided object grounding and robot grasping.
\newblock \emph{arXiv preprint arXiv:2409.19457}, 2024.

\bibitem[Zhou et~al.(2022)Zhou, Yang, Loy, and Liu]{zhou2022CoOp}
Kaiyang Zhou, Jingkang Yang, Chen~Change Loy, and Ziwei Liu.
\newblock Learning to prompt for vision-language models.
\newblock \emph{IJCV}, 130\penalty0 (9):\penalty0 2337--2348, 2022.

\end{thebibliography}
\end{document}